\documentclass[runningheads]{llncs}

\usepackage{eccv}

\usepackage[dvipsnames]{xcolor}

\usepackage{graphicx}
\usepackage{amsmath}
\usepackage{amssymb}
\usepackage{booktabs}
\usepackage{bm}

\usepackage{multirow}
\usepackage{verbatim}
\usepackage{comment}
\usepackage{adjustbox}
\usepackage{wrapfig}

\usepackage{url}            
\usepackage{amsfonts}       
\usepackage{nicefrac}       
\usepackage{microtype}      
\usepackage{xcolor}         

\usepackage{array}
\usepackage{xspace}

\authorrunning{J. Li et al.}

\newcolumntype{L}[1]{>{\raggedright\let\newline\\\arraybackslash\hspace{0pt}}m{#1}}
\newcolumntype{C}[1]{>{\centering\let\newline\\\arraybackslash\hspace{0pt}}m{#1}}
\newcolumntype{R}[1]{>{\raggedleft\let\newline\\\arraybackslash\hspace{0pt}}m{#1}}

\newcommand{\ignore}[1]{}

\renewcommand*{\thefootnote}{\fnsymbol{footnote}}

\makeatletter
\DeclareRobustCommand\onedot{\futurelet\@let@token\@onedot}
\def\@onedot{\ifx\@let@token.\else.\null\fi\xspace}

\makeatother

\definecolor{MyDarkBlue}{rgb}{0,0.08,1}
\definecolor{MyDarkGreen}{rgb}{0.02,0.6,0.02}
\definecolor{MyDarkRed}{rgb}{0.8,0.02,0.02}
\definecolor{MyDarkOrange}{rgb}{0.40,0.2,0.02}
\definecolor{MyPurple}{RGB}{111,0,255}
\definecolor{MyRed}{rgb}{1.0,0.0,0.0}
\definecolor{MyGold}{rgb}{0.75,0.6,0.12}
\definecolor{MyDarkgray}{rgb}{0.66, 0.66, 0.66}

\definecolor{turquoise}{cmyk}{0.65,0,0.1,0.3}

\usepackage{eccvabbrv}

\usepackage{graphicx}
\usepackage{booktabs}

\usepackage[accsupp]{axessibility}  

\usepackage{hyperref}

\usepackage{orcidlink}

\begin{document}
\title{Controllable Human-Object Interaction Synthesis}

\author{Jiaman Li\inst{1}, Alexander Clegg\inst{2}, Roozbeh Mottaghi\inst{2}, \\
Jiajun Wu\inst{1}, Xavier Puig\inst{2}$^\dagger$, C. Karen  Liu\inst{1}$^\dagger$
}
\institute{$^1$Stanford University, $^2$FAIR, Meta}

\maketitle

\renewcommand{\thefootnote}{\fnsymbol{footnote}}
\setcounter{footnote}{2} 

\makeatletter
\def\blfootnote{\xdef\@thefnmark{}\@footnotetext}
\makeatother
\blfootnote{$\dagger$ indicates equal contribution.}

\begin{abstract}
Synthesizing semantic-aware, long-horizon, human-object interaction is critical to simulate realistic human behaviors. In this work, we address the challenging problem of generating synchronized object motion and human motion guided by language descriptions in 3D scenes. 
We propose \textbf{C}ontrollable \textbf{H}uman-\textbf{O}bject \textbf{I}nteraction \textbf{S}ynthesis (CHOIS), an approach that generates object motion and human motion simultaneously using a conditional diffusion model given a language description, initial object and human states, and sparse object waypoints. Here, language descriptions inform style and intent, and waypoints, which can be effectively extracted from high-level planning, ground the motion in the scene. 
Naively applying a diffusion model fails to predict object motion aligned with the input waypoints; it also cannot ensure the realism of interactions that require precise hand-object and human-floor contact.
To overcome these problems, we introduce an object geometry loss as additional supervision to improve the matching between generated object motion and input object waypoints; we also design guidance terms to enforce contact constraints during the sampling process of the trained diffusion model. 
We demonstrate that our learned interaction module can synthesize realistic human-object interactions, adhering to provided textual descriptions and sparse waypoint conditions. Additionally, our module seamlessly integrates with a path planning module, enabling the generation of long-term interactions in 3D environments. Please refer to our \href{https://lijiaman.github.io/projects/chois/}{\textcolor{magenta}{project page}} for the qualitative results.
\end{abstract}  
\section{Introduction}
\label{sec:intro}
Synthesizing human behaviors in 3D environments is critical for various applications in computer graphics, embodied AI, and robotics. Humans effortlessly navigate and engage within their surroundings, performing a plethora of tasks routinely. For example, drawing a chair closer to a desk to create a workspace, adjusting a floor lamp to cast the perfect glow, or neatly storing a suitcase. Each of these tasks requires precise coordination between the human, the object, and the surroundings. These tasks are also deeply rooted in purpose. Language serves as a powerful tool to articulate and convey these intentions. Synthesizing realistic human and object motion guided by language and scene context is the cornerstone of building advanced AI systems that simulate continuous human behaviors in diverse 3D environments.

\begin{figure*}[t!]
\begin{center}
    \centering
    \includegraphics[width=\textwidth]{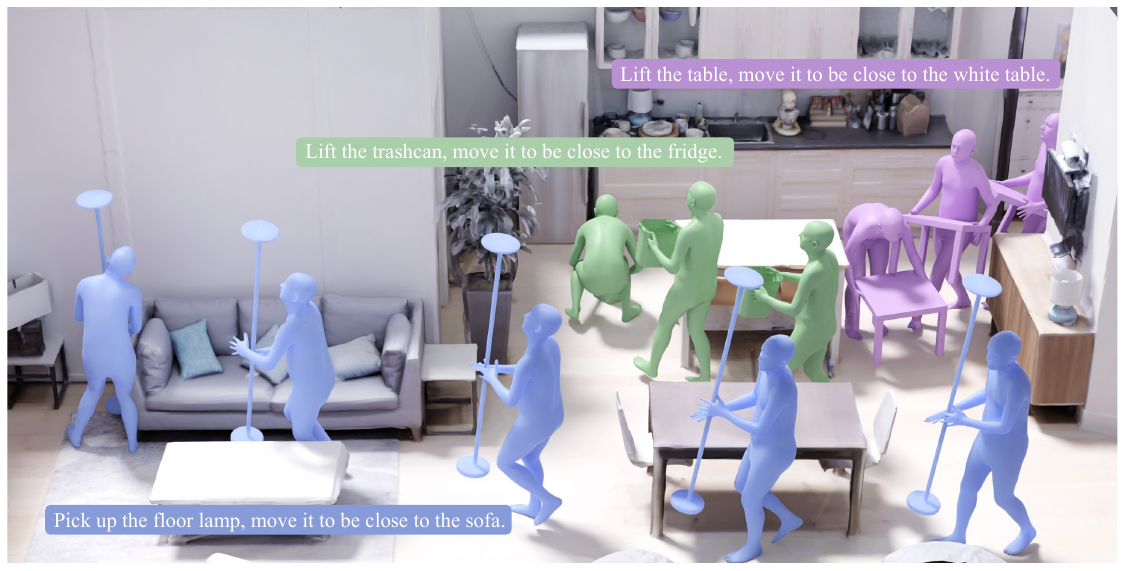}
    \vspace{-6mm}
    \captionof{figure}{Given an initial object and human state, a language description, and sparse object waypoints in a 3D scene, CHOIS generates synchronized object motion and human motion at the same time. }
    \label{figure1}
    \vspace{-10mm}
\end{center}%
\end{figure*}

Although some existing works study the problem of human-scene interaction~\cite{hassan_samp_2021}, they are restricted to scenarios with static objects, such as sitting on a chair, neglecting the highly dynamic interactions that occur frequently in daily life. Recent advances have been made in modeling dynamic human-object interactions, yet these approaches focus solely on smaller objects~\cite{ghosh2022imos,li2023task} or lack the ability to manipulate diverse objects~\cite{hassan2023synthesizing,xie2023hierarchical}. The most recent work on manipulation of larger, diverse objects relies on sequences of past interaction states or complete sequences of object motion~\cite{wan2022learn,xu2023interdiff,li2023object}, thus being incapable of synthesizing both object motion and human motion from initial states alone. Unlike these existing methods, we focus on synthesizing realistic human-object interactions for diverse objects in 3D environments from language and initial states. This problem is challenging primarily for two reasons. First, we need to generate realistic and synchronized motions for both objects and humans. Human hands should maintain appropriate contact with objects during interaction, and object motion should maintain a causal relationship to human actions. Second, 3D scenes are often cluttered with numerous objects, constraining the space of feasible motion trajectories. Thus, the method should accommodate environment clutter, rather than operating under the assumption of an empty scene.

To address these challenges, we leverage waypoints to guide the synthesis process. Starting with a language description outlining the desired human actions and an initial object and human state, we first extract a set of waypoints from the environment. Our goal is thus to generate motions for both humans and objects that align with the directives specified by language, while also conforming to the environmental constraints defined by waypoint conditions derived from 3D scene geometry.

To achieve this, we employ a conditional diffusion model to generate synchronized object and human motion simultaneously, conditioned on language descriptions, initial states, and sparse object waypoints. However, naively applying a diffusion model fails to generate object motions that precisely adhere to the input object waypoints. Additionally, the generated interactions often exhibit issues such as unrealistic contact, foot floating, and objects penetrating the floor. To improve the accuracy of the predicted object motion, we incorporate an object geometry loss during training. Furthermore, we devise guidance terms applied during the sampling process to explicitly enforce contact constraints, thereby directly enhancing the realism of the generated interactions. 
We demonstrate the effectiveness of our learned interaction synthesis module within a system that produces continuous, realistic, and context-aware interactions given language descriptions and 3D scenes.  

To summarize, our work makes the following contributions. First, we identify that the combination of language and object waypoints provides precise and expressive information for human-object interaction synthesis. We show that object waypoints do not need to be dense, which allows us to utilize existing path planning algorithms to generate sparse waypoints that represent long-horizon interactions in complex scenarios. Second, based on this finding, we devise a method that synthesizes human-object interaction guided by language and sparse waypoints of the object, using a conditional diffusion model. Third, we demonstrate that our approach synthesizes realistic interactions on the FullBodyManipulation dataset~\cite{li2023object} and generalizes to novel objects from 3D-FUTURE~\cite{fu20213d}. We also integrate our method into a pipeline that synthesizes long-horizon environment-aware human-object interactions from 3D scenes and language input. 

\section{Related Work}
\label{sec:related}
{\bf{Motion Synthesis from Language.}}
With the development of large-scale high-quality motion capture datasets like AMASS~\cite{AMASS}, there has been a growing interest in generative human motion modeling. BABEL~\cite{punnakkal2021babel} and HumanML3D~\cite{Guo_2022_CVPR} further introduce action labels and language descriptions to enrich the mocap dataset, enabling the development of action-conditioned motion synthesis~\cite{petrovich2021action} and text-conditioned motion synthesis~\cite{Guo_2022_CVPR,petrovich2022temos,tevet2022motionclip}. Prior work has shown that VAE formulation is effective in generating diverse human motion from text~\cite{Guo_2022_CVPR,guo2022tm2t}. Recently, with the success of the diffusion model in this domain~\cite{chen2023executing,barquero2023belfusion,huang2023diffusion,raab2023single,shafir2023human,yuan2023physdiff,zhang2023tedi,tseng2022edge,li2023ego,li2023object,shi2023controllable}, extensive work has explored generating motion from text using conditioning~\cite{tevet2022human,dabral2022mofusion,zhang2022motiondiffuse,karunratanakul2023gmd}. In this work, we also take language descriptions as input to guide our generation. Instead of synthesizing human motion alone, we generate both object motion and human motion conditioned on the text.

\noindent{\bf{Motion Synthesis in 3D Scenes.}}
With the advent of paired scene-motion data~\cite{prox,araujo2023circle,wang2022humanise,HPS,zheng2022gimo} and paired object-motion data~\cite{zhang2022couch,hassan_samp_2021}, approaches~\cite{wang2021synthesizing,wang2022towards,araujo2023circle,zhang2022couch,hassan_samp_2021,kulkarni2023nifty} have been developed to generate human interactions such as sitting on a chair and reaching a target position in 3D scenes. To populate human-object interactions without training on paired scene-motion data, path planning algorithms have been deployed to generate collision-free paths which then guide the human motion generation~\cite{hassan_samp_2021,mir2023generating,zhang2022wanderings,zhao2023synthesizing}. Another line of work leverages reinforcement learning frameworks to train scene-aware policies for synthesizing navigation and interaction motions in static 3D scenes~\cite{lee2023locomotion,xiao2023unified}. In this work, instead of focusing on static scenes or objects, we synthesize interactions with dynamic objects. Also, inspired by approaches that decompose scene-aware motion generation into path planning and goal-guided generation phases, we design an interaction synthesis module conditioned on sparse object waypoints that can be effectively integrated into a scene-aware synthesis pipeline. 

\noindent{\bf{Interaction Synthesis.}}
The field of modeling dynamic human-object interactions has largely focused on hand motion synthesis~\cite{zhang2021manipnet,christen2022d,zheng2023cams}. Recently, with the advent of full-body motion datasets with hand-object interactions~\cite{GRAB:2020,fan2023arctic}, models~\cite{taheri2022goal,wu2022saga} have been developed to synthesize full-body motions preceding object grasping. Some recent studies predict object motion based on human movements~\cite{petrov2023object}, and others~\cite{ghosh2022imos,li2023task,braun2023physically} have taken this further by synthesizing both body and hand motion, subsequently applying optimization to predict object motion.  However, these approaches focus on smaller objects where hand motion is the primary focus.
In terms of manipulating larger objects, some methods train reinforcement learning policies to synthesize box lifting and moving behaviors~\cite{hassan2023synthesizing,xie2023hierarchical,merel2020catch}, yet these models struggle to generalize to manipulation of diverse objects. Based on paired human-object motion data~\cite{bhatnagar22behave,wan2022learn,li2023object}, recent works predict interactions from a sequence of past interaction states~\cite{wan2022learn,xu2023interdiff} or an object motion sequence~\cite{li2023object}, incapable of synthesizing interactions in 3D scenes solely from initial states. In this work, we generate synchronized object and human motion conditioned on sparse object waypoints, serving to ground the resulting trajectories in 3D scenes.   

\noindent{\bf{Concurrent Work.}}
Our work is concurrent with CG-HOI~\cite{diller2023cg} and HOI-Diff~\cite{peng2023hoi}, which use the BEHAVE dataset~\cite{bhatnagar22behave} to synthesize both human motion and object motion from text. Our work also aims to synthesize human-object interactions but differs from the concurrent approaches by integrating textual conditions with sparse waypoint conditions. This unique combination enables our interaction synthesis module to be integrated with path planning modules, enabling long-term interactions in 3D environments. Additionally, we leverage the FullBodyManipulation dataset~\cite{li2023object}, specifically designed for interaction synthesis, offering superior data scale and motion quality compared to BEHAVE~\cite{bhatnagar22behave}.

\section{Method}

\begin{figure*}[t!]
\begin{center}
\includegraphics[width=\textwidth]{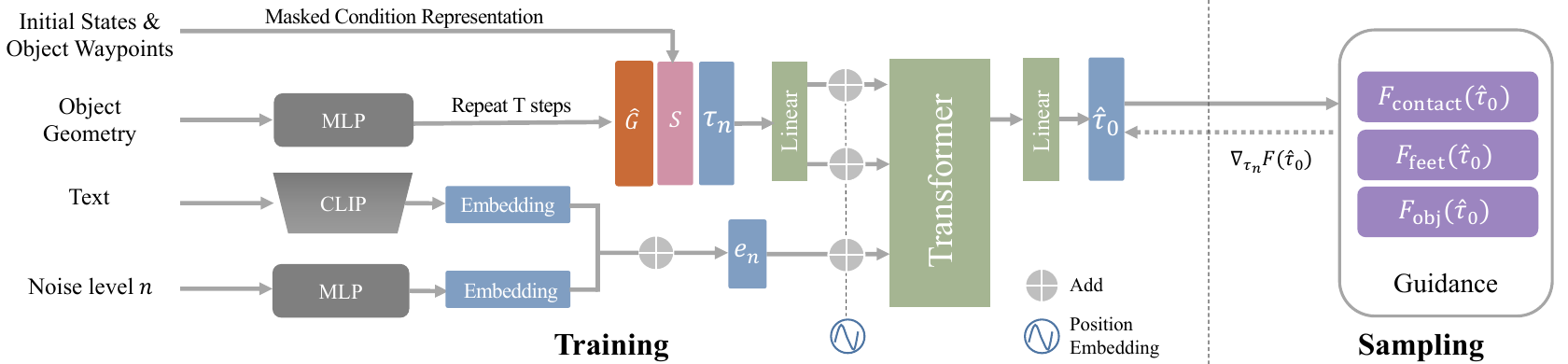}
\end{center}
\vspace{-6mm}
\caption{\textbf{Method Overview.} Given an object geometry, we use the BPS representation to encode the geometry and an MLP to project the features into a low-dimensional vector. This feature vector is concatenated with masked pose states to form conditions for the denoising network. During sampling, we use analytical functions to compute gradients and perturb the generation to satisfy our defined constraints. }
\label{fig:overview}
\vspace{-4mm}
\end{figure*}

Our goal is to generate synchronized object and human motion, conditioned on a language description, object geometry, initial object and human states, and sparse object waypoints. Two primary challenges arise in this context: first, modeling the complexity of synchronized object and human motion while also respecting the sparse condition signals; and second, ensuring the realism of contact between the human and object. To tackle the generation problem of complex interactions, we employ a conditional diffusion model to generate object motion and human motion at the same time. However, naively learning a conditional diffusion model to generate both object motion and human motion cannot ensure the precise contact between hand and object and the realism of the interaction. Thus, we incorporate several constraints as guidance during the sampling process of our trained diffusion model. We illustrate our approach in Figure~\ref{fig:overview}.

\subsection{Data Representation}
\noindent \textbf{Object and Human Motion Representation.}
We denote the human motion as $\bm{X} \in \mathbb{R}^{T \times D}$, where $T$ and $D$ represent the time steps and dimension of the human pose. $\bm{X}_t$, corresponding to the human pose at frame $t$, consists of global joint positions and 6D continuous rotations~\cite{zhou2019continuity}. 
We adopt the widely used parametric human model, SMPL-X~\cite{smplx} to reconstruct the human mesh from the pose and shape parameters. To represent the object motion, we use two components: the global 3D position and the relative rotation.
The global position is represented by the centroid of the object, while the relative rotation, denoted as  $\bm{R}_\text{rel}$ at frame $t$,  is expressed with respect to the input object's geometry $\bm{V}$ such that $\bm{V}_t = \bm{R}_\text{rel}\bm{V}$, where $\bm{V}_t$ represent the vertices of object at frame $t$. We denote the object motion by $\bm{O} \in \mathbb{R}^{T \times 12}$.

\noindent \textbf{Object Geometry Representation.}
We represent the object geometry using the Basis Point Set (BPS) representation~\cite{prokudin2019efficient}. Following prior work~\cite{li2023object}, we begin by sampling a set of basis points from the volume of a ball with a 1-meter radius. Subsequently, for each sampled point, we calculate the minimum Euclidean distance to the nearest point on the object's mesh. Alongside this, we record the directional vectors from the basis points to their nearest neighbors. The resulting BPS representation is denoted as $\bm{G} \in \mathbb{R}^{1024 \times 3}$, representing 1024 sampled points each with a vector indicating their spatial relationship to the object's surface.

\noindent \textbf{Input Condition Representation.}
 We first use an MLP to project the object BPS representation $\bm{G}$ to a low-dimensional vector which is then broadcasted to each frame denoted as $\hat{\bm{G}} \in \mathbb{R}^{T \times 256}$ following~\cite{li2023object}. We then adopt a masked motion data representation denoted as $\bm{S} \in \mathbb{R}^{T \times (12 + D)}$ to represent the initial states and waypoint conditions. The initial state contains the human pose and object pose at the first frame.
 The waypoint conditions consist of a series of 2D object positions for every 30 frames, and a 3D object position at the final frame. The remainder of $\bm{S}$ is padded with zeros. The encoded object geometry vector and the masked motion condition vector are then concatenated, serving as part of the input for our denoising network. For effectively integrating language conditions, we utilize CLIP~\cite{radford2021learning} as a text encoder to extract language embeddings.

\subsection{Interaction Synthesis Model} 
\noindent \textbf{Conditional Diffusion Model.}
We utilize a conditional diffusion model~\cite{ho2020denoising} to generate synchronized object and human motion. To improve the realism of hand-object interaction, our model also predicts contact labels $\bm{H} \in \mathbb{R}^{T \times 2}$ for both the left and right hands. These predicted contact labels play a crucial role in guiding the sampling process, ensuring more accurate and realistic hand-object contacts in the generated motion sequence. The complete data representation in our model is denoted as $\bm{\tau} = \{\bm{X}, \bm{O}, \bm{H}\}$, encapsulating motion and contact data.

The conditional signals of our model, denoted as $\bm{c}$, include initial states, sparse object waypoints, the object BPS representation, and language descriptions. The diffusion model consists of a forward diffusion process that progressively adds noise to the clean data $\bm{\tau}_{0}$ and a reverse diffusion process which is trained to reverse this process. The forward diffusion process introduces noise for $N$ steps formulated using a Markov chain,
\begin{align}
\label{eq:forward_diffusion_step}
    q(\bm{\tau}_{n}|\bm{\tau}_{n-1}) := \mathcal{N}(\bm{\tau}_{n}; \sqrt{1-\beta_{n}}\bm{\tau}_{n-1}, \beta_{n}\bm{I}), \\
     q(\bm{\tau}_{1:N}|\bm{\tau}_{0}) := \prod_{n=1}^{N} q(\bm{\tau}_{n}|\bm{\tau}_{n-1}),
\end{align}
where $\beta_{n}$ represents a fixed variance schedule and $\bm{I}$ is an identity matrix. Our goal is to learn a model $p_{\theta}$ to reverse the forward diffusion process, 
\begin{equation}
\label{eq:reverse_diffusion_step}
    p_{\theta}(\bm{\tau}_{n-1}|\bm{\tau}_{n}, \bm{c}) := \mathcal{N}(\bm{\tau}_{n-1}; \bm{\mu}_{\theta}(\bm{\tau}_n, n, \bm{c}), \bm{\Sigma}_{n}),
\end{equation}
where $\bm{\mu}_{\theta}$ denotes the predicted mean and $\bm{\Sigma}_{n}$ is a fixed variance. 
Learning the mean can be re-parameterized as learning to predict the clean data representation $\bm{\tau}_0$. The objective~\cite{ho2020denoising} is defined as
\begin{equation}
\label{eq:loss}
     \mathcal{L} = \mathbb{E}_{\bm{\tau}_0, n}||\hat{\bm{\tau}}_{\theta}(\bm{x}_{n}, n, \bm{c}) - \bm{\tau}_{0}||_{1}.
     \vspace{-2mm}
\end{equation}

\noindent \textbf{Model Architecture.}
 We employ a transformer architecture~\cite{transformer} as our denoising network. Our input consists of object geometry $\hat{\bm{G}}$, masked motion conditions $\bm{S}$, and noisy data representation $\bm{\tau}_{n}$ at noise level $n$. The input is projected to a sequence of feature vectors using a linear layer. We employ an MLP to embed the noise level $n$. Then we combine the noise level embedding and the language embedding to form a single embedding vector denoted as $\bm{e}_{n}$. The embedding vector $\bm{e}_{n}$ has the same dimension as these feature vectors and is fed to the transformer along with these vectors. The final prediction $\hat{\bm{\tau}}_{0}$ is made by projecting the updated feature vectors of the transformer excluding the time step corresponding to $\bm{e}_{n}$. The interaction synthesis model is illustrated in Figure~\ref{fig:overview}. 

\noindent \textbf{Object Geometry Loss.}
During the training phase, we incorporate an additional loss to improve the object motion prediction. Utilizing the Basis Point Set (BPS) representation, we initially compute the nearest neighbor points on the object mesh in rest pose for each of the fixed set of points. From these, we sample 100 points out of the 1024 nearest neighbors to capture a rough outline of the object's shape. These selected points are defined as $\bm{K}_\text{rest} \in \mathbb{R}^{100 \times 3}$, representing our selected object vertices at rest pose.

At each time step in our model, the predicted object rotation (converted to relative rotation with respect to the object geometry in rest pose) and position are employed to calculate the corresponding positions of these selected vertices. This is represented by the following equation, where $\hat{\bm{R}}_t$ and $\hat{\bm{d}}_{t}$ denote the predicted rotation and translation of the object, and $\bm{K}_{t}$ refers to the ground truth vertices at time step $t$. The object geometry loss is computed as
\begin{equation}
\label{eq:loss_k}
     \mathcal{L}_{\text{obj}} = \sum_{t=1}^{T}||\hat{\bm{R}_t}\bm{K}_\text{rest} + \hat{\bm{d}}_t - \bm{K}_t||_{1}.
\end{equation}
This loss function plays a critical role in guiding the model to accurately predict the transformation of the object.

\subsection{Guidance}
During the training phase of our interaction synthesis model, there are no explicit contact constraints enforced in the losses. Incorporating loss terms such as hand-object contact loss, and object-floor penetration loss poses a challenge for training. First, these types of loss terms are computationally expensive and would slow down training significantly. Second, introducing more loss terms requires meticulously balancing different losses which usually necessitates re-training models with different settings. Instead, enforcing these constraints during test time is more flexible and makes it easier to select appropriate weights for different terms. Thus, to refine our generated interactions, we propose the application of guidance during the sampling process.

In the diffusion model framework, classifier guidance is commonly applied during test time to control the generation process in order to satisfy specific objectives or constraints. A typical approach to applying classifier guidance~\cite{dhariwal2021diffusion} is to perturb the noisy predicted mean at each denoising step. This is formulated as $\Tilde{\bm{\mu}} = \bm{\mu} - \alpha\bm{\Sigma}_{n}\nabla_{\bm{\mu}}F(\bm{\mu})$, where $\bm{\mu}$ denotes the predicted mean at denoising step $n$ defined by Equation~\ref{eq:reverse_diffusion_step}. $F$ represents a learned or analytical function that determines how much the predicted mean should be penalized and $\alpha$ represents the strength of the perturbation. This guidance computes the gradient with respect to the noisy mean, requiring $F$ to be trained on noisy data or a deterministic function designed for noisy data. Another approach is reconstruction guidance~\cite{ho2022video}, which has proven to be effective for controlling the generation process in prior work~\cite{rempe2023trace,karunratanakul2023gmd,kulkarni2023nifty}. Instead of perturbing the noisy mean, it perturbs the predicted clean data representation $\hat{\bm{\tau}}_{0}$ using the gradient with respect to the noisy input data representation $\bm{\tau}_{n}$. The process is formally represented as
\begin{equation}
\label{eq:guidance}
     \Tilde{\bm{\tau}}_{0} = \hat{\bm{\tau}}_{0} - \alpha\bm{\Sigma}_{n}\nabla_{\bm{\tau}_n} F(\hat{\bm{\tau}}_{0}).
\end{equation}

In this work, we leverage reconstruction guidance~\cite{ho2022video} in the sampling process as we empirically found it to be more stable. We define multiple analytical functions as guidance terms which we will introduce in the following sections. 

\noindent \textbf{Hand-Object Contact Guidance.}
We have implemented a specialized contact guidance function to improve the hand-object contact accuracy for frames generated by our model. This function is specifically designed to address cases where a noticeable distance exists between the hands and the object, thereby improving the realism of the interaction.
The contact guidance function is defined as follows:
\begin{equation}
\label{eq:loss_contact}
     F_{\text{contact}} = \left\|\bm{M}_{l} \odot \left|\bm{J}_{l} - \bm{V}_{l}\right|\right\|_1 + \left\|\bm{M}_{r} \odot \left|\bm{J}_{r} - \bm{V}_{r}\right|\right\|_1.
\end{equation}

$\bm{M}_{l}$ and $\bm{M}_{r}$ are binary masks for the left and right hand, respectively. These masks are derived from the predicted contact labels $\bm{H}$, with $\bm{M}_{l}, \bm{M}_{r} \in \mathbb{R}^{T \times 1}$ and are defined as $\bm{M}_{l}, \bm{M}_{r} = \left(\bm{H} > 0.95\right)$. This thresholding identifies frames where contact is likely to occur. The symbol $\odot$ represents the Hadamard product, applying these masks element-wise to the absolute differences between the hand positions ($\bm{J}_{l}$, $\bm{J}_{r} \in \mathbb{R}^{T \times 3}$) and nearest points on the object mesh ($\bm{V}_{l}$, $\bm{V}_{r} \in \mathbb{R}^{T \times 3}$). 


\noindent \textbf{Feet-Floor Contact Guidance.}
When generating joint positions and rotations, our model operates without awareness of the body's shape. Consequently, using the SMPL-X model~\cite{smplx} with predicted root positions, joint rotations, and a test subject's specific body shape parameters to reconstruct the human mesh can sometimes lead to scenarios where the feet do not touch the floor. To rectify this, we implement a guidance term that encourages realistic feet-floor contact.

The joint positions of the left and right toes are represented as $\bm{J}_{l}$ and $\bm{J}_{r}$, respectively. We identify the supporting foot in each frame by comparing the z components of these two joints at each frame. We also introduce a threshold height $h = 0.02$ meters, which is determined from the analysis of foot height in the ground truth motion. The guidance term is defined as follows:
\begin{equation}
\label{eq:loss_feet_floor}
     F_{\text{feet}} = ||min(\bm{J}_{l}^{z}, \bm{J}_{r}^{z}) - h||_{2}.
\end{equation}
This function computes the norm of the vertical difference between the lowest point of either toe and the threshold height $h$. 

\noindent \textbf{Object-Floor Penetration Guidance.}
To address the issue of generated object states potentially penetrating the floor, we integrate an additional guidance function into the sampling process. Given that our floor is positioned at the plane where $z = 0$, we define the guidance term as follows:
\begin{equation}
\label{eq:loss_object_floor}
     F_{\text{obj}} = ||min(\bm{V}^z, 0)||_{1},
\end{equation}
where $\bm{V}^z$ represents the z-coordinate of the object vertices.

During inference, we apply multiple guidance concurrently defined as follows, 
\begin{equation}
\label{eq:loss_all_guidance}
     F_{\text{all}} = \lambda_{1}F_{\text{contact}} + \lambda_{2}F_{\text{feet}} + \lambda_{3}F_{\text{obj}},
\end{equation}
where $\lambda_{1}$, $\lambda_{2}$, $\lambda_{3}$ denote the loss weights. We apply the guidance in the last 10 denoising steps only since the prediction in the early steps is extremely noisy.   

\section{Experiments}

We first introduce the datasets and evaluation metrics. Then we show comparisons of our proposed approach against the baselines. We further conduct a human perceptual study to complement our evaluation and ablation study to verify the effectiveness of our proposed guidance terms. Moreover, we demonstrate an application that generates long-term interactions conditioned on object waypoints extracted from 3D scenes. 

\subsection{Datasets}
\noindent\textbf{The FullBodyManipulation dataset}~\cite{li2023object} consists of 10 hours of high-quality, paired object and human motion, including interaction with 15 different objects.  However, our study does not encompass the generation of motion for articulated objects, leading us to exclude sequences related to two such objects (vacuum and mop). We employ this dataset both for training our interaction model and for evaluating the generated results. The training set comprises 15 subjects, with an additional 2 subjects designated for testing, adhering to the dataset partitioning used in OMOMO~\cite{li2023object}. We specifically chose the FullBodyManipulation dataset~\cite{li2023object} over BEHAVE~\cite{bhatnagar22behave} due to several limitations in the latter. BEHAVE is not tailored for interaction synthesis, presenting challenges such as noticeable jittery motions, limited data scale and a lack of locomotion.

\noindent\textbf{The 3D-FUTURE dataset}~\cite{fu20213d} includes 3D models of various furniture items. From this dataset, we select 17 objects representing diverse types (such as chairs, tables, floor lamps, and boxes). This dataset serves to test our model's ability to generalize to objects it has not previously encountered. Given that the 3D-FUTURE dataset only includes 3D models, we integrate these objects with motion from the testing set of the FullBodyManipulation dataset~\cite{li2023object} to generate input conditions for evaluation. In particular, given an object in 3D-FUTURE, we take a motion from FullBodyManipulation belonging to the same object category, and extract the 2D coordinates for the object positions every 30 frames to represent the input waypoints. 

\subsection{Evaluation Metrics}

\noindent\textbf{Condition Matching Metric:} This metric calculates the Euclidean distance between the predicted and input object waypoints. It includes the start and end position errors $\left(T_{s}, T_{e}\right)$, and waypoint errors $\left(T_{xy}\right)$ measured in centimeters (cm).

\noindent\textbf{Human Motion Quality Metric:} This metric encompasses the foot sliding score (FS), foot height $\left(H_{feet}\right)$, \textit{Fr\'{e}chet Inception Distance} $\left(FID\right)$ and \textit{R-precision} $\left(R_{prec}\right)$. FS is the weighted average of accumulated translation in the xy plane, following prior work~\cite{he2022nemf}, measured in centimeters (cm). $H_{feet}$ assesses the height of the feet, also in centimeters. $R_{prec}$ and $FID$ are computed following the text-to-motion task~\cite{Guo_2022_CVPR}. $R_{prec}$ (top-3) measures whether the generated motion is consistent with the text. $FID$ assesses the motion quality by computing the discrepancy between the distributions of ground truth and generated motions. 

\noindent\textbf{Interaction Quality Metric:} This metric assesses the accuracy of hand-object interactions, encompassing both contacts and penetrations. For contact accuracy, it employs precision $\left(C_{prec}\right)$, recall $\left(C_{rec}\right)$, and F1 score $\left(C_{F_1}\right)$ metrics following prior work~\cite{li2023object}. Additionally, it includes contact percentage $\left(C_{\%}\right)$, determined by the proportion of frames where contact is detected. To compute the penetration score $\left(P_{hand}\right)$, each vertex of the hand $V_i$ is used to query the precomputed object's Signed Distance Field (SDF). This process yields a corresponding distance value $d_i$ for each vertex. The penetration score is then derived by computing the average of the negative distance values (representing penetration), formalized as $\frac{1}{n}\sum_{i=1}^{n}|min(d_i, 0)|$, measured in centimeters (cm). 

\noindent\textbf{Ground Truth (GT) Difference Metric:} This metric measures the deviation of generated results from the ground truth motion. It comprises the mean per-joint position error (MPJPE), translation error of the root joint $\left(T_{root}\right)$, and object position error $\left(T_{obj}\right)$, all computed using the Euclidean distance between the predicted and actual ground truth positions in centimeters (cm). Additionally, this metric includes the root joint orientation error $\left(O_{root}\right)$ and the object orientation error $\left(O_{obj}\right)$. These errors are calculated with the Frobenius norm of the rotational difference, formulated as $||\bm{R}_{pred}\bm{R}_{gt}^{-1} - \bm{I}||_{2}$ where $\bm{R}_{pred}$ and $\bm{R}_{gt}$ represent the predicted and ground truth rotation matrices respectively.

\subsection{Results}
\noindent \textbf{Baselines.}
As there is no prior work presenting a solution for our task, we adapt related works such as InterDiff~\cite{xu2023interdiff}, MDM~\cite{tevet2022human}, and OMOMO~\cite{li2023object} to fit our problem setting in order to establish baseline comparisons. InterDiff~\cite{xu2023interdiff} focuses on anticipating human-object interactions using the previous 10 frames. MDM~\cite{tevet2022human} generates human motion from language descriptions. OMOMO~\cite{li2023object} synthesizes human motion based on provided object motion trajectories. We adapt InterDiff to accept additional input conditions including text and sparse waypoints. For MDM, we update the model to incorporate our object geometry representation and sparse waypoints. Additionally, we enhance MDM to include our object motion representation as an additional output. OMOMO requires a sequence of object states to generate full-body human poses; therefore, we implement a linear interpolation strategy for object positions based on the provided start and end positions, as well as predefined waypoints in the xy-plane, while maintaining consistent object rotation from the initial frame throughout the sequence. Furthermore, we introduce two variations, Pred-OMOMO and GT-OMOMO, as part of our ablation studies. Pred-OMOMO combines our text-conditioned object motion synthesis module with OMOMO. GT-OMOMO utilizes ground truth object motion as input for OMOMO.
Additionally, we evaluate our approach CHOIS against two ablations: CHOIS w/o $L_\text{obj}$ and CHOIS w/o $F_\text{all}$. CHOIS w/o $L_\text{obj}$ is trained as a conditional diffusion model but does not include an additional object geometry loss. This variant allows us to understand the baseline performance of the diffusion model in a straightforward setup. In contrast, CHOIS w/o $F_\text{all}$ incorporates the object geometry loss in its training process but operates without guidance during inference. This approach lets us explore the effectiveness of object geometry loss during training while assessing the model's capability in the absence of guidance.

\begin{table*}[t!]
\small
\caption{\textbf{Interation synthesis} on the FullBodyManipulation dataset~\cite{li2023object}.} 
    \label{tab:single_window_cmp_seen}
    \vspace{-2mm}
    \centering 
\footnotesize{
\setlength{\tabcolsep}{1pt}
  \resizebox{\textwidth}{!}{ 
\begin{tabular}{lcccccccccccccccc} 
 \toprule 
 & \multicolumn{3}{c}{Condition Matching} & \multicolumn{4}{c}{Human Motion} & \multicolumn{5}{c}{Interaction} & \multicolumn{4}{c}{GT Difference}  \\
 \cmidrule(lr){2-4}\cmidrule(lr){5-8}\cmidrule(lr){9-13}\cmidrule(lr){14-17}  
 Method     & $T_{s}\downarrow$ & $T_{e}\downarrow$ & $T_{xy}\downarrow$ & $H_{feet}\downarrow$ & FS$\downarrow$  & $R_{prec}\uparrow$ & $FID\downarrow$ & $C_{prec}\uparrow$ & $C_{rec}\uparrow$ & $C_{F_1}\uparrow$ & $C_{\%}$ & $P_{hand}\downarrow$ & MPJPE$\downarrow$  & $T_{root}\downarrow$  & $T_{obj}\downarrow$ & $O_{obj}\downarrow$   \\
        \midrule
         Interdiff~\cite{xu2023interdiff} & 0.00 & 158.84 & 72.72 & \textbf{0.90} & 0.42 & 0.08 & 208.0 & 0.63 & 0.28 & 0.33 & 0.27 & 0.55 & 25.91 & 63.44 & 88.35 & 1.65   \\
        MDM~\cite{tevet2022human} & 5.18 & 33.07 & 19.42 & 6.72 & 0.48 & 0.51 & 6.16 & 0.72 & 0.47 & 0.53 & 0.43 & 0.66 & 17.86 & 34.16 & 24.46 & 1.85   \\
        \midrule 
        Lin-OMOMO~\cite{li2023object} & 0.00 & 0.00 & 0.00 & 7.21 & 0.41 & 0.29 & 15.33 & 0.68 & 0.56 & 0.57 & 0.54 & \textbf{0.51} & 21.73 & 36.62 & 17.12 & 1.21 \\
        Pred-OMOMO~\cite{li2023object}  &  2.39 & 8.03 & 4.15 & 7.08 & 0.40 & 0.54 & 4.19 & 0.73 & 0.66 & 0.66 & 0.62 & 0.58 & 18.66 & 28.39 & 16.36 & 1.05  \\
        GT-OMOMO~\cite{li2023object}  & 0.00 & 0.00 & 0.00 & 7.10 & 0.41 & 0.48 & 5.69 & 0.77 & \textbf{0.66} & 0.67 & 0.59 & 0.55 & 15.82 & 24.75 & 0.00 & 0.00   \\ 
        \midrule
        CHOIS w/o $L_{obj}$ & 5.76 & 14.16 & 8.44 & 6.55 & 0.40 & 0.65 & 3.26 & 0.75 & 0.50 & 0.55 & 0.43 & 0.66 & \textbf{14.34} &  \textbf{21.97} & 15.53 & \textbf{0.98} \\
        CHOIS w/o $F_\text{all}$  & 1.75 & 6.61 & \textbf{2.69} & 6.64 & 0.38 & \textbf{0.65} & 3.58 & 0.78 & 0.49 & 0.55 & 0.41 & 0.65 & 15.23 & 24.13 & \textbf{11.51} & 0.99 \\
        CHOIS (ours) & \textbf{1.71} & \textbf{6.31} & 2.87 & \textbf{4.20} & \textbf{0.35} & 0.64 & \textbf{0.69} & \textbf{0.80} & 0.64 & \textbf{0.67} & 0.54 & 0.59 & 15.30 &  24.43 & 12.53 & 0.99 \\
        \bottomrule
\end{tabular}
}
}
\vspace{-2mm}
\end{table*}

\begin{table}[!t]
 \caption{\textbf{Interaction synthesis} on the 3D-FUTURE dataset~\cite{fu20213d}. }
    \label{tab:single_window_cmp_unseen}
    \centering
    \vspace{-2mm}
    \begin{adjustbox}{width=0.7\textwidth} {
    \small
    \footnotesize{
    \setlength{\tabcolsep}{2pt}
      \begin{tabular}{lccccccccccc}
        \toprule

        & \multicolumn{3}{c}{Condition Matching} & \multicolumn{4}{c}{Human Motion} & \multicolumn{2}{c}{Interaction} \\
 \cmidrule(lr){2-4}\cmidrule(lr){5-8}\cmidrule(lr){9-10}
        
        & $T_{s}\downarrow$ & $T_{e}\downarrow$ & $T_{xy}\downarrow$  & $H_{feet}\downarrow$ & FS$\downarrow$ & $R_{prec}\uparrow$ & $FID\downarrow$ & $C_{\%}$ & $P_{hand}\downarrow$   \\
        
        \midrule
        InterDiff~\cite{xu2023interdiff} & 0 & 161.26 & 72.77 & -0.26 & 0.42 & 0.09 & 207.3 & 0.24 & \textbf{0.11} \\
        MDM~\cite{tevet2022human} & 12.58 & 40.55 & 28.72 & 7.02 & 0.49 & 0.53 & 8.50 & 0.34 & 0.26 \\
        \midrule 
        Lin-OMOMO~\cite{li2023object} & 0 & 0 & 0 & 6.32 & 0.42 & 0.23 & 23.17 & 0.44 & 0.11 \\
        Pred-OMOMO~\cite{li2023object} & 4.15 & 9.03 & 3.89 & 6.08 & 0.40 & 0.46 & 3.74 & 0.50 & 0.18 \\
        \midrule 
        CHOIS w/o $L_{obj}$  & 6.70 & 13.73 & 7.99 & 5.68 & 0.41 & \textbf{0.66} & 3.26 & 0.36 & 0.30 \\
        CHOIS w/o $F_\text{all}$ & 5.75 & 7.96 & \textbf{2.68} & 5.84 & 0.39 & 0.62 & 4.78 & 0.33 & 0.26 \\
        CHOIS (ours) & \textbf{4.12} & \textbf{7.35} & 2.92 & \textbf{3.75} & \textbf{0.38} & 0.62 & \textbf{1.60} & 0.48 & 0.15 \\
        \bottomrule
       
      \end{tabular}
    }
    }\end{adjustbox}
    \vspace{-2mm}
\end{table}

\noindent \textbf{Results on the FullBodyManipulation Dataset.}
We evaluate our approach using objects from the FullBodyManipulation dataset~\cite{li2023object} as shown in Table~\ref{tab:single_window_cmp_seen}. Introducing object geometry loss notably improves the condition matching metric. Furthermore, adding guidance during inference leads to better contact accuracy, reduced hand-object penetration, and less foot floating. 

InterDiff cannot adequately adhere to the input waypoints and text since the input conditions are entangled. The condition embedding, which includes the past 10 frames, point cloud features, text, and sparse waypoints, is summed up to predict future frames. This approach leads to suboptimal performance in condition matching metrics and $R_{prec}$. Also, we observe feet-floor penetration issues in InterDiff's generated results, resulting in a lower foot height. MDM can synthesize plausible interactions but, as seen in the Interaction metrics, struggles with generating realistic contacts since it does not enforce any contact constraints.

\begin{figure*}[t!]
\begin{center}
\includegraphics[width=0.9\textwidth]{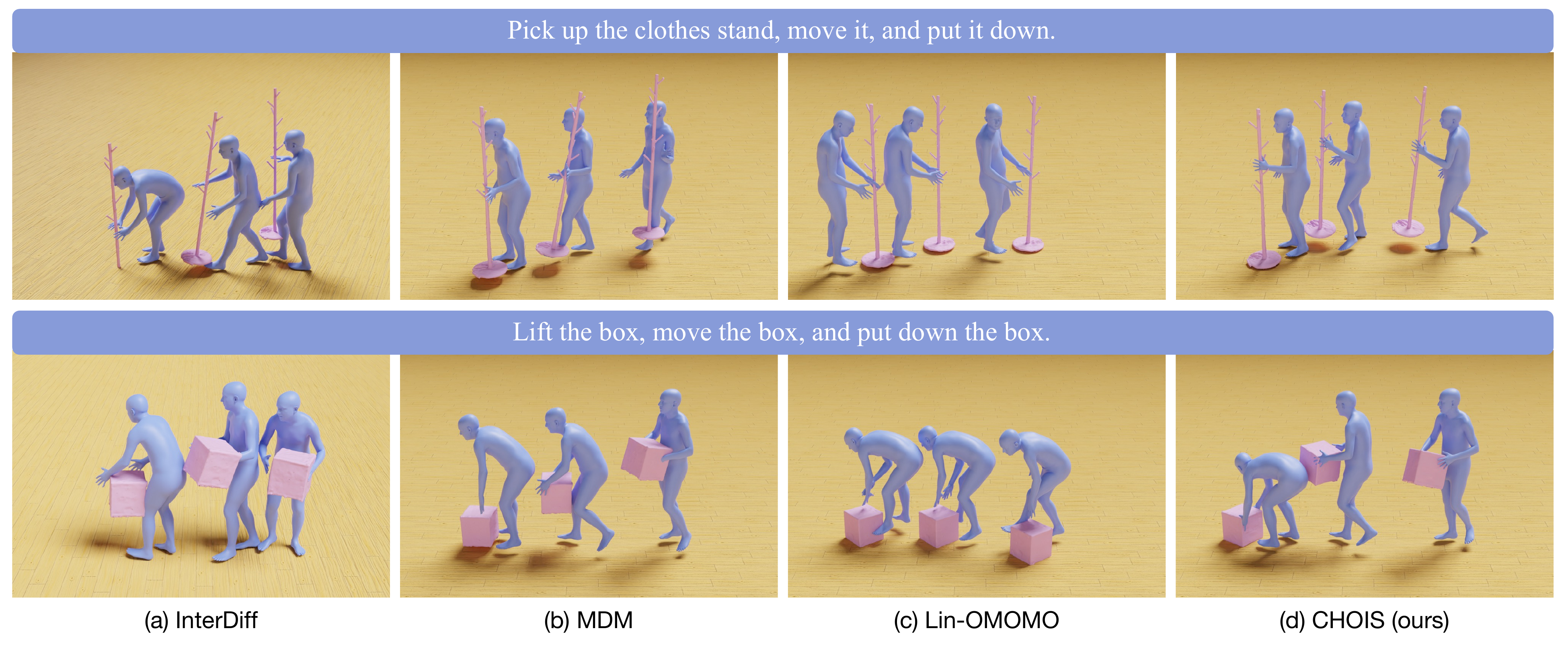}
\end{center}
\vspace{-6mm}
\caption{\textbf{Qualitative results} of the FullBodyManipulation dataset~\cite{li2023object}.}
\label{fig:qualitative_res}
\vspace{-2mm}
\end{figure*}

Lin-OMOMO shows zero deviation from the input object trajectory as it only predicts human motion and does not alter the object motion input at the sparse input locations. Pred-OMOMO demonstrates improved contact metrics compared to the baselines but is still inferior to our CHOIS in terms of condition matching and human motion quality. Moreover, Pred-OMOMO requires three stages during inference, one from our object motion synthesis module and two from OMOMO, whereas CHOIS operates as a single-stage model. GT-OMOMO, requiring a ground truth object motion sequence for input, shows comparable performance in interaction and GT difference metrics. However, the motions it generates suffer from foot floating issues, leading to a larger $H_{feet}$ and $FID$. 
We also showcase qualitative comparisons against the baselines in Figure~\ref{fig:qualitative_res}. 

\noindent \textbf{Results on the 3D-FUTURE Dataset.}
To test our model's ability to generalize to new objects, we conduct evaluations using the 3D-FUTURE dataset~\cite{fu20213d}. As shown in Table~\ref{tab:single_window_cmp_unseen}, our proposed method outperforms the baselines. 

\begin{figure}[t!]
\begin{center}
\includegraphics[width=0.85\linewidth]{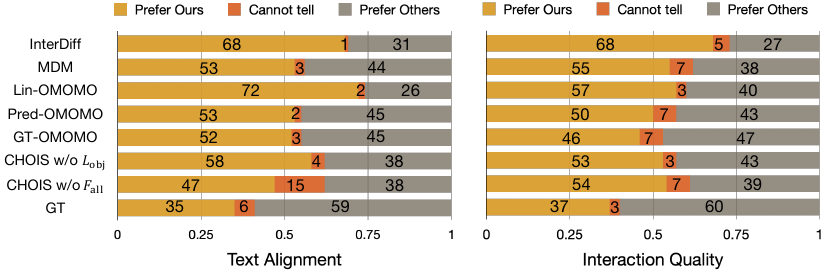}
\end{center}
\vspace{-6mm}
  \caption{\textbf{Results of human perceptual studies.} The numbers shown in the chart represent the percentage (\%) over motion preferences.}
  \label{fig:human_study}
\vspace{-2mm}
\end{figure}

\begin{table*}[t!]
\small
\caption{\textbf{Ablation study} on the FullBodyManipulation dataset~\cite{li2023object}. We measure the effect of different guidance terms in the human and object motion generation.} 
    \label{tab:ablation_study_seen}
\vspace{-2mm}
\centering
\resizebox{\textwidth}{!}{
\footnotesize{
\setlength{\tabcolsep}{1pt}
\begin{tabular}{lcccccccccccccccc} 
 \toprule 
 & \multicolumn{3}{c}{Condition Matching} & \multicolumn{4}{c}{Human Motion} & \multicolumn{5}{c}{Interaction} & \multicolumn{4}{c}{GT Difference}  \\
 \cmidrule(lr){2-4}\cmidrule(lr){5-8}\cmidrule(lr){9-13}\cmidrule(lr){14-17}  
 Method     & $T_{s}\downarrow$ & $T_{e}\downarrow$ & $T_{xy}\downarrow$ & $H_{feet}\downarrow$ & FS$\downarrow$ & $R_{prec}\uparrow$ & $FID\downarrow$ & $C_{prec}\uparrow$ & $C_{rec}\uparrow$ & $C_{F_1}\uparrow$ & $C_{\%}$ & $P_{hand}\downarrow$ & MPJPE$\downarrow$  & $T_{root}\downarrow$  & $T_{obj}\downarrow$ & $O_{obj}\downarrow$   \\
        \midrule
       CHOIS w/o $F_\text{contact}$ & \textbf{1.70} & 6.42 & 2.70 & \textbf{3.93} & \textbf{0.32} & \textbf{0.66} & 0.74 & 0.78 & 0.49 & 0.55 & 0.41 & 0.65 & 15.41 & \textbf{23.63} & \textbf{11.44} & 0.99 \\
        CHOIS w/o $F_\text{feet}$ & 1.72 & 6.34 & 2.90 & 6.65 & 0.39 & 0.63 & 3.76 & \textbf{0.81} & \textbf{0.64} & 0.66 & 0.54 & \textbf{0.58} & 15.44 & 25.09 & 13.31 & 0.99 \\
        CHOIS w/o $F_\text{all}$  & 1.75 & 6.61 & \textbf{2.69} & 6.64 & 0.38 & 0.65 & 3.58 & 0.78 & 0.49 & 0.55 & 0.41 & 0.65 & \textbf{15.23} & 24.13 & 11.51 & 0.99 \\
        CHOIS (ours) & 1.71 & \textbf{6.31} & 2.87 & 4.20 & 0.35 & 0.64 & \textbf{0.69} & 0.80 & 0.64 & \textbf{0.67} & 0.54 & 0.59 & 15.30 &  24.43 & 12.53 & 0.99 \\
        \bottomrule
\end{tabular}
}
}
\vspace{-2mm}
\end{table*}

\begin{figure*}[t!]
\begin{center}
\includegraphics[width=\textwidth]{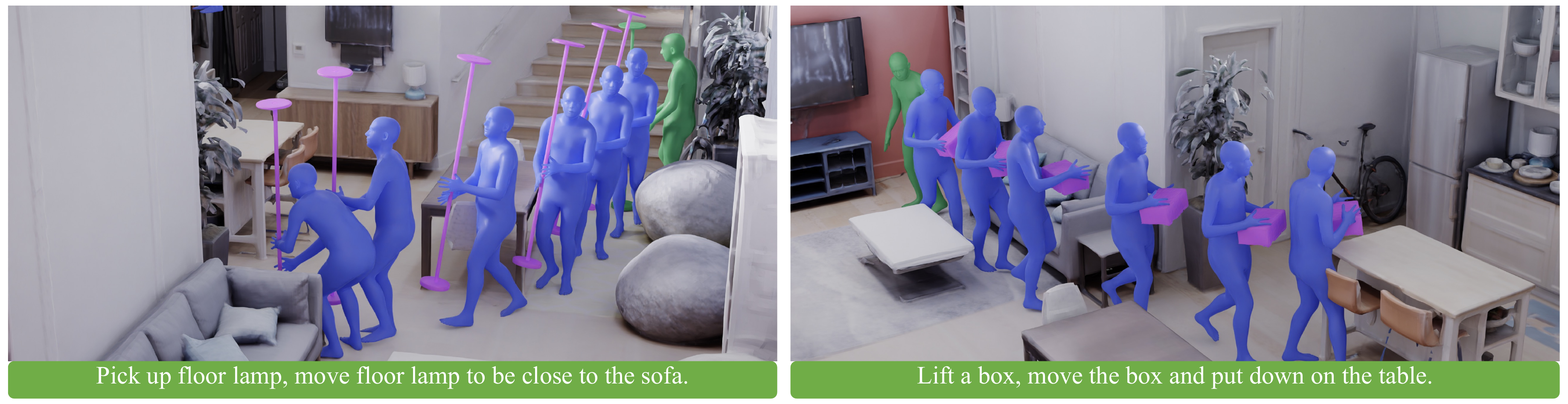}
\end{center}
\vspace{-6mm}
\caption{\textbf{Long-term interaction synthesis.} Given language descriptions, a 3D scene with semantic labels, and initial human and object states, we synthesize long-term human-object interactions. The initial state is shown in green.}
\label{fig:qualitative_res_w_scene}
\vspace{-1mm}
\end{figure*}

\noindent \textbf{Human Perceptual Study.}
We conduct two human perceptual studies to further complement the evaluation of our approach. The first study assesses the consistency between the generated interactions and the text input. The second study evaluates the overall quality of these generated interactions. For each of these studies, we generate 100 sequences using each method, including baselines, OMOMO ablations, our CHOIS model, our own ablations, and the ground truth. This results in a set of 800 pairs. We employ Amazon Mechanical Turk (AMT) for evaluation. Each sequence pair is reviewed by 10 different AMT workers. The results are illustrated in Figure~\ref{fig:human_study}. 

\begin{table}[!t]
 \caption{\textbf{Long-term interaction synthesis results} on the FullBodyManipulation~\cite{li2023object} and 3D-FUTURE datasets~\cite{fu20213d}. $^{\ast}$ represents the results on the 3D-FUTURE dataset. 
     }
    \label{tab:long_seq_cmp}
    \centering
    \vspace{-2mm}
    \begin{adjustbox}{width=0.7\textwidth} {
    \small
    \footnotesize{
    \setlength{\tabcolsep}{2pt}
      \begin{tabular}{lccccccccc}
        \toprule

        & \multicolumn{3}{c}{Condition Matching} & \multicolumn{2}{c}{Human Motion} & \multicolumn{2}{c}{Interaction} \\
 \cmidrule(lr){2-4}\cmidrule(lr){5-6}\cmidrule(lr){7-8}
        
        & $T_{s}\downarrow$ & $T_{e}\downarrow$ & $T_{xy}\downarrow$  & $H_{feet}\downarrow$ & FS$\downarrow$ & $C_{\%}$ & $P_{hand}\downarrow$   \\
        
        \midrule
        
        CHOIS w/o $F_\text{all}$ & \textbf{1.50} & \textbf{7.19} & \textbf{5.10} & 6.01 & \textbf{0.43} & 0.49 & 0.70  \\
        CHOIS & 2.22 & 9.94 & 5.73 & \textbf{4.57} & 0.46 & \textbf{0.63} & \textbf{0.69} \\
        \midrule
        CHOIS$^{\ast}$ w/o $F_\text{all}$ & 6.29 & \textbf{9.39} & \textbf{5.26} & 4.77 & \textbf{0.39} & 0.42 & 0.55  \\
        CHOIS$^{\ast}$ & \textbf{5.62} & 12.08 & 5.95 & \textbf{4.27} & 0.41 & \textbf{0.65} & \textbf{0.32}  \\
        \bottomrule
       
      \end{tabular}
    }
    }\end{adjustbox}
    \vspace{-2mm}
\end{table}

\subsection{Ablation Study}
We conduct an ablation study to validate the effectiveness of our proposed guidance terms. As shown in Table~\ref{tab:ablation_study_seen}, our hand-object contact guidance and feet-floor contact guidance are both critical. Without the hand-object contact guidance, the contact percentage degrades obviously. Without the feet-floor contact guidance, the height of the feet increases indicating there exists severe foot floating issues. We are not ablating object-floor penetration guidance as object-floor penetration issues are not common and this term is primarily designed for preventing penetration artifacts in qualitative results.    

\subsection{Application}

\begin{figure}[t!]
\begin{center}
\includegraphics[width=0.9\linewidth]{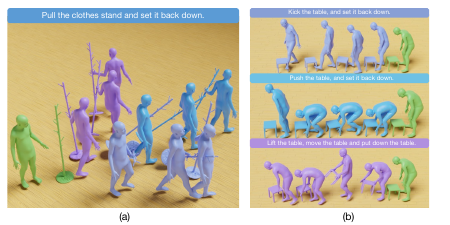}
\end{center}
\vspace{-6mm}
  \caption{\textbf{Results of interaction synthesis} using the same text but different waypoints (a) and using the same waypoints but different text (b). The initial state is in green.}
  \label{fig:same_text_diverse_waypoints}
\vspace{-5mm}
\end{figure}
This section presents a practical application of our method, enabling the synthesis of human-object interactions within 3D scenes, driven by language descriptions. We utilize 3D scenes from the Replica Dataset~\cite{replica}. The process begins by composing language descriptions that specify the desired interactions, identifying both the objects involved and their intended positions. For example, the language description can be ``pull the floor lamp to be close to a shelf''. We also define a set of primitive functions used to sample target 3D positions from 3D scenes. This set includes functions like sampling points on an object's surface or near it. GPT-3~\cite{brown2020language} is used to extract key information including the interaction object and target objects, and to select the appropriate primitive functions from our predefined function set. Combining the information with the semantic labels of the scene point cloud, we can determine the target 3D positions.

We leverage Habitat~\cite{habitat19iccv, szot2021habitat} to generate collision-free paths within the scene given the start and target object positions. However, as Habitat provides waypoints without corresponding time steps, we need to adapt these to our learned module. We apply heuristics to create waypoints at fixed intervals of 30 frames, which serve as the input conditions for our 
model. The text input for our learned module excludes the directional component (e.g., “Pick up the floor lamp and move it”), focusing solely on the action and object. An example of this application is shown in Figure~\ref{fig:qualitative_res_w_scene}, demonstrating how our learned interaction synthesis model 
effectively synthesizes human-object motion following a description in a 3D scene.
Table~\ref{tab:long_seq_cmp} includes a quantitative evaluation of the generated motion.
In addition, we showcase the results using the same text input but different waypoints and the results using the same waypoints but different text in Figure~\ref{fig:same_text_diverse_waypoints}, demonstrating the effectiveness of the control using object waypoints and text.

\section{Conclusion}
In conclusion, our work addresses the problem of human-object interaction synthesis conditioned on language descriptions and sparse object waypoints. By employing a conditional diffusion model, we successfully generate object and human motions that are not only synchronized but also resonate with given language descriptions. We incorporate object geometry loss during training which significantly improves the performance of object motion generation. We also propose effective guidance terms used during the sampling process which enhance the realism of the generated results. Moreover, we demonstrate that our learned interaction module can be integrated into a pipeline that synthesizes long-term interactions given language and 3D scenes.  

\noindent \textbf{Acknowledgments.} This work is in part supported by the Wu Tsai Human Performance Alliance at Stanford University, the Stanford Institute for Human-Centered AI (HAI), NSF CCRI \#2120095, ONR MURI N00014-22-1-2740, and Meta. Part of the research was done during Jiaman Li's internship at FAIR, Meta.

\bibliographystyle{splncs04}
\bibliography{main}

\end{document}